\documentclass[conference]{IEEEtran}
\IEEEoverridecommandlockouts
\usepackage{cite}
\usepackage{amsmath,amssymb,amsfonts}
\usepackage{algpseudocode}
\usepackage{algorithm}
\usepackage{graphicx}
\usepackage{textcomp}
\usepackage{xcolor}
\usepackage{xspace}
 \usepackage{kotex}
\usepackage{soul}
\usepackage{booktabs}
\usepackage{subfig}
\usepackage{graphicx}

\begin{document}
\title{Spatio-Temporal Split Learning for Autonomous Aerial Surveillance using Urban Air Mobility (UAM) Networks}


\author{
\IEEEauthorblockN{Yoo Jeong Ha}
\IEEEauthorblockA{\textit{Korea University}\\
Seoul, Korea \\
annaha17@korea.ac.kr}
\and
\IEEEauthorblockN{Soyi Jung}
\IEEEauthorblockA{\textit{Hallym University} \\
Chuncheon, Korea \\
sjungs@hallym.ac.kr}
\and
\IEEEauthorblockN{Jae-Hyun Kim}
\IEEEauthorblockA{\textit{Ajou University} \\
Suwon, Korea \\
jkim@ajou.ac.kr}
\and
\IEEEauthorblockN{Marco Levorato}
\IEEEauthorblockA{\textit{UC-Irvine} \\
California, USA \\
levorato@uci.edu}
\and
\IEEEauthorblockN{Joongheon Kim}
\IEEEauthorblockA{\textit{Korea University}\\
Seoul, Korea \\
joongheon@korea.ac.kr}
}

\maketitle

\begin{abstract}

Autonomous surveillance unmanned aerial vehicles (UAVs) are deployed to observe the streets of the city for any suspicious activities. This paper utilizes surveillance UAVs for the purpose of detecting the presence of a fire in the streets. An extensive database is collected from UAV surveillance drones. With the aid of artificial intelligence (AI), fire stations can swiftly identify the presence of a fire emerging in the neighborhood. Spatio-temporal split learning is applied to this scenario to preserve privacy and globally train a fire classification model. Fires are hazardous natural disasters that can spread very quickly. Swift identification of fire is required to deploy firefighters to the scene. In order to do this, strong communication between the UAV and the central server where the deep learning process occurs is required. Improving communication resilience is integral to enhancing a safe experience on the roads. Therefore, this paper explores the adequate number of clients and data ratios for split learning in this UAV setting, as well as the required network infrastructure.

\end{abstract}

\section{Introduction}
The compounding effects of climate change have led to unforeseeable climate-related shifts in the ecosystem. Fires are the epitome of the vicious cycle of climate change. Bushfires are the leading causes of elevated greenhouse gas emissions; yet, ironically, the effects of global warming, such as increased air temperature and dried-out trees, which act as the fuel for a fire, create the perfect climate condition for wildfire to occur. A study~\cite{mietkiewicz2020line} shows that humans caused 97\,\% of the wildfires that threatened homes in the wildlife between 1992 and 2015. Most fire departments deploy their firefighters only after the fire has spread to the point where it takes hours or even days and countless firemen and women to extinguish a flame completely. A method that determines the presence of ignition before it starts spreading can save valuable resources. Hence, the necessity of surveillance UAVs that accurately detect the presence of a fire in the streets will is becoming ever more paramount~\cite{tvt201905shin,tvt202106jung}. 


Street-level surveillance is becoming a popular tool to fight off arson or larceny; it can be used as the next-level fire detection apparatus. An increasing amount of people are utilizing personal surveillance gadgets to protect their safety. Personal-level closed-circuit television (CCTV), such as smart doorbells, have features to ensure customers' privacy and security~\cite{9406452}. Yet, these domestic surveillance devices may include technology that pries on the privacy of its neighbors. In other words, these cameras may contain sensitive personal information such as car plate numbers, or people entering their homes, which must be protected. Aside from advancing capabilities of surveillance technologies, a method to ensure a comprehensive and appropriate system that protects sensitive information captured through surveillance devices is essential. Yet, no privacy protecting steps are involved in such surveillance devices. 

Therefore, this paper creates a scenario in which multiple surveillance UAVs observe the city for signs of ignition without exposing classified information captured by the UAVs. This is achieved by learning the fire classification model through a split learning process with the street data captured by the surveillance UAVs. Split learning is a method that splits the deep neural network into two groups: a client and a server~\cite{vepakomma2018split, poirot2019split,matsubara2021split, ha2021spatio}. Here, the clients will be the surveillance UAVs that capture the data of the ignitions in the streets; and the centralized server is the platform where the heavy classification neural network (CNN) is trained. Each of the participating UAVs locally runs their captured images up to the first hidden layer. The output from the first hidden layer is then transferred to the centralized server~\cite{matsubara2019distilled, matsubara2020head}. 
\begin{figure*}[t!]
    \begin{center}
        \includegraphics[width=0.7\linewidth]{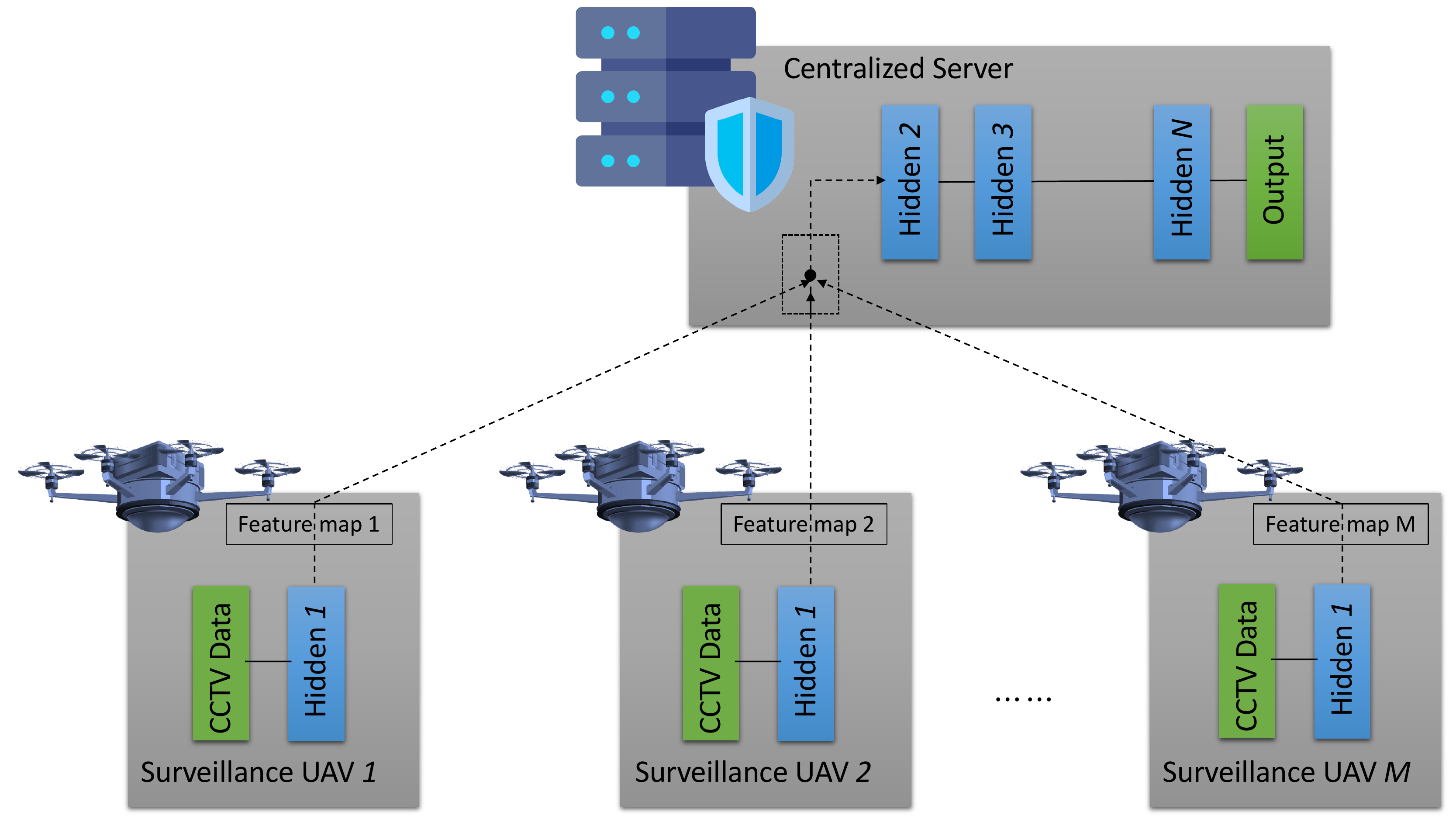}
    \end{center}
    \caption{The surveillance UAV split learning framework.}
    \label{fig:framework}
\end{figure*}

By physically splitting the deep learning process and allowing the feature map, not the original data, to be exposed to the external network, the sensitive information obtained from the UAV data is protected. Moreover, the UAV is a mobile device that has battery and energy limitations. Thus rather than a completely local computation, a split learning system that offloads the most computational high learning to a server is introduced in this paper.


This paper examines how split learning involving multiple UAVs resolves the issue of data-imbalance and protects any sensitive information captured by surveillance UAVs. Since the performance of split learning is established, this paper experiments with altering the number of clients and the split ratio in a surveillance UAV environment. Hence in this study, we explore the effect of increasing the number of UAVs on the classification accuracy of split learning and the impact of varying the data split ratio.

For split learning to perform at its highest potential, a robust network structure is required. For the output from the UAV to be sent to a centralized server with little latency, the network must be stable. A large channel capacity is needed to offload the encrypted feature map to decrease any communication-related delay components. Furthermore, surveillance UAVs are energy-limited mobile devices in which UAV caching {\cite{video2, 9512492, twc201912choi,9473012,tmc202106malik}} and fast data transmission {\cite{8119812,7045578,7961189,tvt2021jung}} with streaming quality adaptation~\cite{ton201608kim,jsac201806choi} to the server are important factors. Hence, uploading surveillance footage to the cloud {\cite{tmc201907koo,7748468,6410040,video1, mm2017koo, video3}} in a wireless transmission setting must also to considered. 

Thus, this paper combines two main ideas: identifying the most optimal split ratio and the most effective number of UAV for the highest performance level.   

The main contributions to this paper are as follows: 
\begin{itemize}
    \item We test the impact of changing the number of UAVs on the fire classification accuracy. This study analyzes the realistic scenario where multiple UAVs would observe the suburb for unnatural behavior. Hence, we analyze the effects of increasing the number of number of UAVs on split learning performance. 
    \item We identify the impact of varying the data split ratio between the surveillance UAVs on the performance of the fire classification model. As it will be explored in
    Section~\ref{sec:systemmodel}, the data-imbalance problem is a widely experienced issue in the deep neural network field. Experimenting with the amount of data each UAV holds in our proposed split learning algorithm, will provide an interesting insight into the matter of data-imbalance. 
    \item We determine the most optimal number of UAVs and the most optimal data split ratio for split learning. These variables are altered to find the most effective setting for our proposed method of splitting the classification model between the UAVs and a server.
    
\end{itemize}

The rest of the paper is organized as follows. Section~\ref{sec:systemmodel} gives the problem formulation that is to be explored in this study. It delves into the network model the UAVs are connected by, and the overall system model of the surveillance UAV split learning. Section~\ref{framework} highlights the entire UAV framework by looking at the pseudo-code for the split learning algorithm. The experimental setup and performance is explored in Section~\ref{result}. Then, the paper concludes with Section~\ref{sec:sec5}.

\section{System Model}\label{sec:systemmodel}
\subsection{System Model: Surveillance UAV Split Learning Architecture}

Our proposed system model consists of surveillance UAVs, denoted as clients, and a centralized server. We assume that multiple UAVs are connected to a nearby server via a mmWave channel. These surveillance UAVs roam around a city, recording data of the streets. These data are used to create a global model that successfully classifies the data as images with and without a presence of a fire. As emphasized throughout this paper, these UAVs are battery and energy-limited mobile devices that do not have the capacity to build an entire deep learning model on their own. Hence, the UAVs engage in the new split learning process where the rest of the heavy computation is conducted in a nearby server. 

The surveillance UAV split learning framework is illustrated in Figure~\ref{fig:framework}. The green layer in Figure~\ref{fig:framework} on the surveillance UAV's end is the data input layer. This data then passes through the one blue hidden layer located locally in the surveillance UAV. The output from the first hidden layer is then sent to a nearby server. The surveillance UAV uses mmWave to communicate with the centralized server using wireless back-haul communication. Hence, each UAV only processes its own data up to the first hidden layer.

The output feature maps from each of the UAVs are then concatenated at the server. By doing this, the resource-limited mobile device offloads its data to a more computationally compatible server. The concatenated feature maps now become the input to the rest of the deep learning model the server trains. 

\subsection{Problem Statement}
Data-imbalance is a severe issue in the field of deep learning that can be solved via split learning. Training a deep neural network with insufficient amounts of data will lead to overfitting. Resolving this issue serves as the number one priority in building a deep learning model. Therefore, this study utilizes split learning to resolve the issue of data imbalance and preserve the privacy of its original data. Besides the privacy preservation aspect, the significant advantage of split learning is that multiple clients can share the knowledge to collaboratively build a global network.

Since numerous clients can contribute their data, the small data size of a particular client does not matter. Even if UAVs with a limited amount of data want to engage in split learning, they can do so without causing the model to become overfitted solely due to the characteristic of split learning itself. By enabling UAVs to cooperate in split learning, regardless of their data size, data-imbalance and overfitting problems can be resolved.
 
To analyze this claim, an experiment varying the size of data in each participating UAV is conducted. For each number of clients, the data split ratios are altered between no data-imbalance, data-imbalance, and extreme data-imbalance. The specific split ratios for the three different client numbers are summarized in Table~\ref{tab:acc}. The split ratio called no data-imbalance refers to the case where all the participating UAVs hold an equal amount of data. This is a non-realistic scenario, and it is only used as a benchmark to compare the results. For an unbiased observation, the accuracy of the equally split data ratio is compared to those from the three different numbers of UAVs. 


\begin{algorithm}[h!]
\begin{algorithmic}[1]
\Statex $\hspace{-1.5em}$ \textbf{Assumtions:}~ Batch size $B$, clients $\emph{C}$, number of clients $\emph{n}$, predicted value ${\hat{y}}$, input data $I$, number of input data ${I_n}$, number of label ${l_a}$, and output convolution layer $O^{l}$, output pooling layer $O^{l_{p}}$.
\vspace{0.4em}

\Procedure{Client}{}\\
\textbf{for} {Client = \{1, $\dotsm$, $n$\}} {s.t. $n$\ = 3, 4, 5} 
\textbf{do}\\
	$\hspace{1.5em}$ \textbf{for} \textit{Training data set} = \{1, $\dotsm$, $x$\} \textbf{do}\\
	$\hspace{3em}$ Calculate Conv. \\
	$\hspace{3em}$ $\triangleright$ $f_{c}$ = $\mathrm{Conv}(O_{client}^{l-1},w^{l},I_n,l_a) =net_{I_n,l_a}^{l}$\\
	$\hspace{3em}$ Calculate activation function. \\ 
	$\hspace{3em}$ $\triangleright$ $f_{R}$ = $\mathrm{Relu}(net_{I_n,l_a}^{l})$ \\
    $\hspace{1.5em}$ $\triangleright$ Send feature $f_{c_{1}},f_{c_{2}},\dotsm,f_{c_{n}}$ to server.\\
    $\hspace{1.5em}$  
    \textbf{end for}\\
    \textbf{end for}
\EndProcedure
 	\caption{Spatio-temporal split learning for UAVs}
	\label{algorithm1}
    \end{algorithmic}
\end{algorithm}

\section{UAV Split Learning Framework}\label{framework}

The CNN model for fire classification is split between the participating UAVs and the centralized server. The distinct split of learning processes between the client and server can be seen in the algorithm pseudo-code. The pseudo-code for the client's learning process is outlined in Algorithm 1. The UAV is called the client in split learning and runs the deep learning model only up to the first hidden layer. The captured images of the neighborhood locally run through a hidden layer that only contains a Conv2D layer, as seen in (line 4) and (line 5). After the Conv2D layer, the image passes through a Relu activation function as seen in (line 6) and (line 7). The mobile UAV is only assigned one layer to conserve the restricted power it has. After locally running the captured street images, the UAV offloads the data to a nearby server, where heavy computation workloads occur. 

The output from the UAV is a distorted feature map, which is an encryption of the original data. The captured street images from a surveillance UAV may contain sensitive details such as the resident's facial and plate number information. The first hidden layer, the step in (line 5) in the UAV, distorts the raw image to the point where the original privacy-sensitive details cannot be referred back to the raw image. The extent to which the privacy-preserving layer protects the original image is shown in Figure~\ref{img:feature}. 

\begin{algorithm} [h!]
\begin{algorithmic}[1]
\Procedure{Server}{}\\
    Receive input data from client : $f_{k} , k=(1,\dotsm,n)$\\
   Concatenate all features $\sum_{k=1}^{n}f_{c}^{k}$\\
\textbf{for} \textit{epoch = 1, $E$} \textbf{do}\\
    $\hspace{1.5em}$ \textbf{for} \textit{Training data set} \textbf{do}\\
        $\hspace{3em}$ Calculate Conv, activation function and Pool. $\vspace{0.3em}$ \\
    	$\hspace{3em}$ $\triangleright$ $f_{c} = \mathrm{Conv}(O^{l-1},w^{l},I_n,l_a) = net_{I_n,l_a}^{l}$\\
    	$\hspace{3em}$ $\triangleright$ $f_{R}$ = $\mathrm{Relu}(net_{I_n,l_a}^{l})$ \\
    	$\hspace{3em}$ $\triangleright$ $f_{p}=\mathrm{Pool}(f_{c},I_n,I_a)$\\
    	Repeat the above steps.\\ 
    	$\hspace{1.5em}$ $\triangleright$ ${\hat{y}}$ is calculated using $O^{l_{p}}$, lines 5, 7, and 20.\\
    $\hspace{1.5em}$ $\triangleright$ Calculate loss. \\
    $\hspace{1.5em}$ $\triangleright$ Update the model: update weights $w$ $\leftarrow$ $w$ $\cdot $ $\alpha$.\\
	    $\hspace{1.5em}$  \textbf{end for}\\
 \textbf{end for}
 \EndProcedure
 	\caption{Spatio-temporal split learning for server}
	\label{algorithm2}
    \end{algorithmic}
\end{algorithm}

This distorted feature map is the only information that is exposed to the external network. The output from the UAV (line 8) in Algorithm 1 is transmitted to a centralized server. The detailed pseudo-code for the server side's computation is outlined in Algorithm 2. Each of the outputs from the client is concatenated at the server's end as denoted in (line 3). The concatenated image becomes the input to the deep learning model, where it trains to classify whether there is a fire in the streets or not. The server contains the rest of the hidden layers of the deep learning model. The hidden layers at the server's side consist of a Conv2D layer and a MaxPooling layer. Hence, (line 6) through to (line 9) is repeated for the number of hidden layers the server holds. The system UAV split learning framework is illustrated in Figure~\ref{img:feature}.

\section{Performance Evaluation}\label{result}

\subsection{Experiment Setup}
In a UAV setting, the individual UAV specifications, such as battery power, can impact data transfer from the local device to the server. Thus, we assume the following conditions for the purpose of this paper.  
\begin{itemize}
    \item All the surveillance UAVs run the CNN model up to the first hidden layer.
    \item The energy, power consumption, and battery status are assumed to be the same for all the UAVs.
\end{itemize}

This study experiments with the effect of varying the number of clients. The impact of increasing the number of participating UAVs from 3 to 4 to 5 is explored. In other words, four individual UAVs will run their data locally up to the first hidden layer and send the output, which is the feature map, to one centralized server. The new concatenated input, which is the concatenation of the four separate feature maps, is then used to train the majority of the classification model at the server's end.

\begin{table}[t!]
\small
\begin{center}
	\begin{tabular}{c||c}
    \toprule[1.0pt]
    \textbf{Parameters}  & \textbf{CCTV Flame}  \\
    \midrule
Epochs              & 50                        \\ 
Loss                & Binary crossentropy       \\ 
Activation Function & Sigmoid                   \\ 
Batch Size          & 32                        \\ 
Input Size          & 64 $\times$ 64 $\times$ 1 \\ 
Model               & Custom CNN with VGG19     \\ 
    \bottomrule[1.0pt]
	\end{tabular}
\end{center}
\caption{The CNN configuration to classify the captured surveillance UAV images.}
\label{tab:cnn}
\end{table}

To conduct the study on the impact of varying the number of UAVs and the data split ratio on the performance of split learning, this paper utilizes flames and non-flame CCTV images from Kaggle\footnote{https://www.kaggle.com/ritupande/fire-detection-from-cctv}. The research on varying the number of participating UAV clients and data split ratio is conducted using 533 non-flame CCTV images and 331 flame images. 
The configuration of the CNN model used to classify the captured surveillance UAV image as those containing flames or not is outlined in Table {\ref{tab:cnn}}. Note that the model setup is the same for a fair experiment for all the three varying numbers of UAVs.

\begin{figure}[t]
    \begin{center}
        \includegraphics[width=0.8\linewidth]{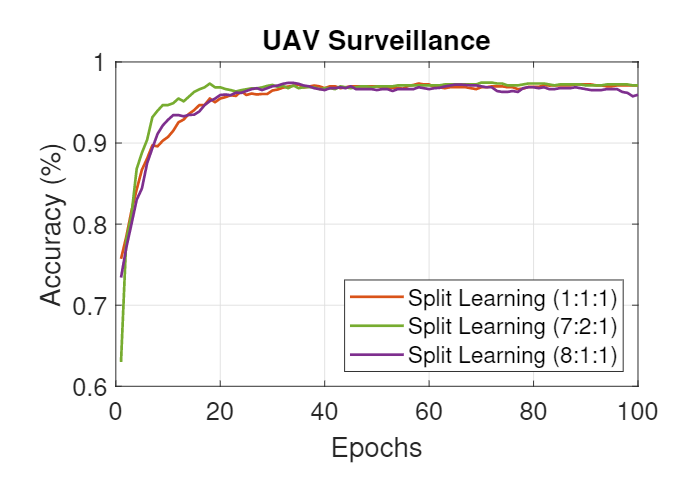}
    \end{center}
    \caption{Classification accuracy with three UAVs.}
    \label{exp:acc3}
\end{figure}

\begin{figure}[t]
    \begin{center}
        \includegraphics[width=0.8\linewidth]{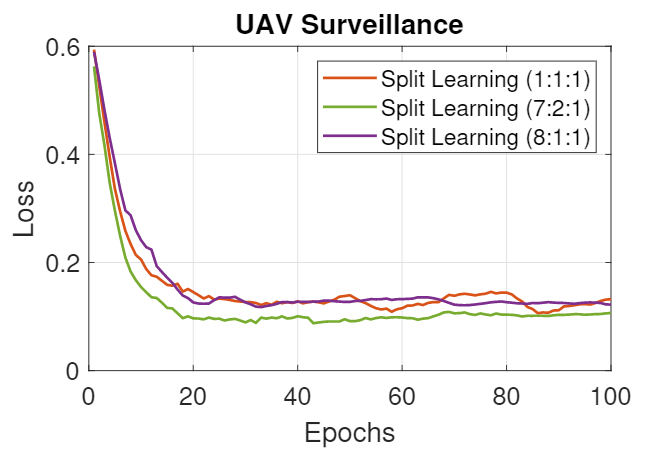}
    \end{center}
    \caption{Loss with three UAVs.}
    \label{exp:loss3}    
    \vspace{-3mm}
\end{figure}

\begin{figure}[t]
    \begin{center}
        \includegraphics[width=0.8\linewidth]{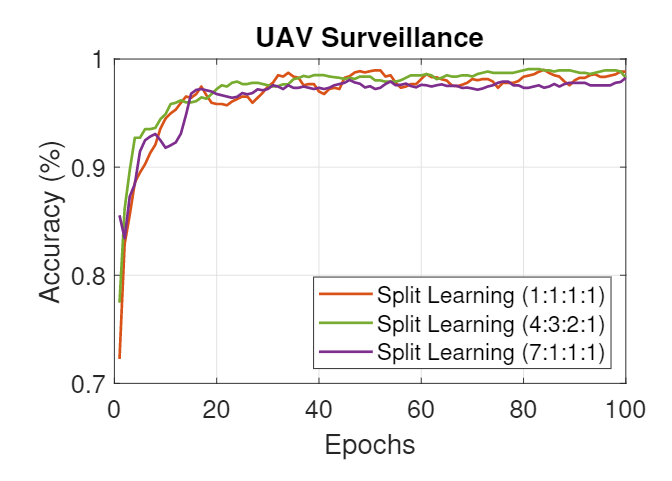}
    \end{center}
    \caption{Classification accuracy with four UAVs.}
    \label{exp:acc4}
\end{figure}

\begin{figure}[t]
    \begin{center}
        \includegraphics[width=0.8\linewidth]{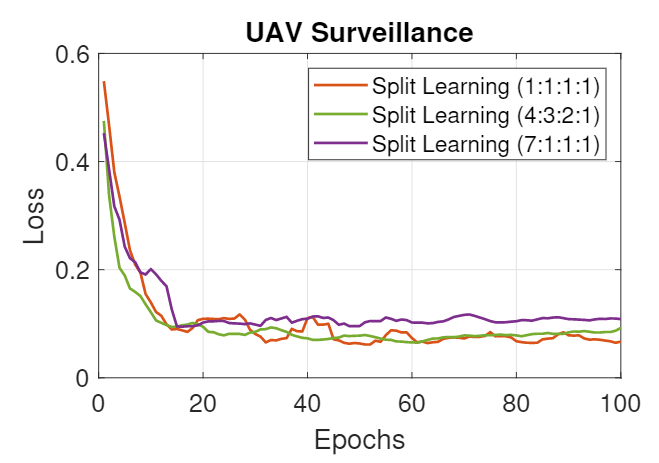}
    \end{center}
    \caption{Loss with four UAVs.}
    \label{exp:loss4}
\end{figure}

\begin{table*}[t]
\normalsize
\begin{center}
    \scalebox{1}{
	\begin{tabular}{r||ccc|ccc|ccc}
    \toprule
     Number of UAVs & & 3 & & & 4 & & & 5 \\
    \midrule [1.0pt]
    Data split ratio   & 1:1:1  & 7:2:1 & 8:1:1 & 1:1:1:1 & 4:3:2:1 & 7:1:1:1 & 1:1:1:1:1 & 4:2:2:1:1 & 6:1:1:1:1\\ 
    Fire classification accuracy (\%)          & 97.0 & 97.1 & 96.5 & 98.4  & 98.8 & 97.6   & 97.7  & 98.8   &  97.0\\  [0.2ex]
    \bottomrule
	\end{tabular}
	}
\caption{Classification accuracy of various number of UAVs and split ratios using surveillance UAV images.}
    \vspace{-3mm}
\label{tab:acc}
\end{center}
\end{table*}

\subsection{Performance Evaluation of Split Learning}
\begin{figure}[t]
    \begin{center}
        \includegraphics[width=0.8\linewidth]{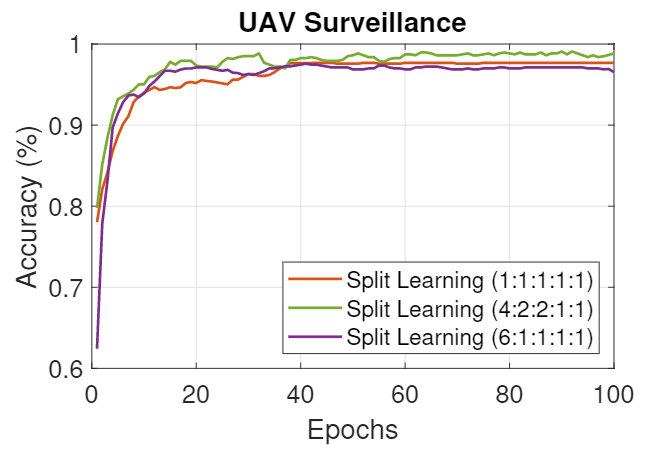}
    \end{center}
    \caption{Classification accuracy with five UAVs.}
    \label{exp:acc5}
\end{figure}

\begin{figure}[t]
    \begin{center}
        \includegraphics[width=0.8\linewidth]{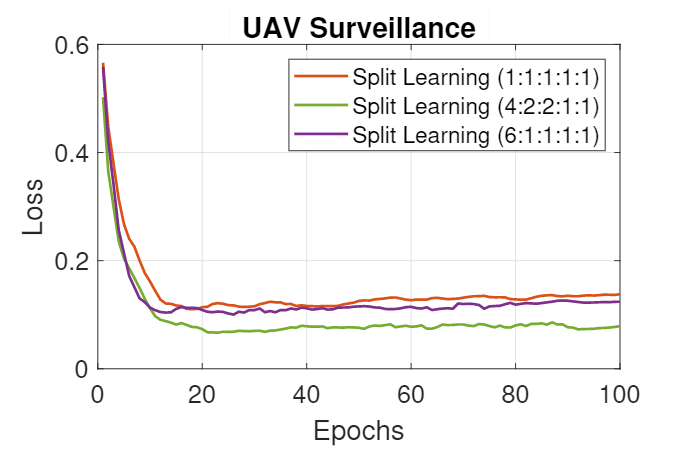}
    \end{center}
    \caption{Loss with five UAVs.}
    \label{exp:loss5}
\end{figure}

This section discusses experimental results for varying the number of UAVs and the split data ratio. The classification accuracy and loss of each of the cases are graphed in Figures~\ref{exp:acc3} to~\ref{exp:loss5}. Figures~\ref{exp:acc3}, \ref{exp:acc4}, and \ref{exp:acc5} show the accuracy rate of 3, 4, and 5 UAVs participating in split learning respectively. Within the three cases, three different data split ratios are applied to test the impact of data-imbalance on classification performance. The exact classification accuracy at which the training converged of all the conducted experiments is summarized in Table~\ref{tab:acc}. Furthermore, it can also be noted that the accuracy level amongst the different data split ratios within the same number of clients does not vary drastically. 
Judging by the precise numbers outlined in Table~\ref{tab:acc}, the highest accuracy is achieved when the data split ratio is has a slight data-imbalance. For each of the cases with 3, 4, and 5 UAVs, the best performance is when the data is split into a 7:2:1, 4:3:2:1, and 4:2:2:1:1, respectively. 

Additionally, examining the loss values for the training of the classification model confirms the most optimal number of UAVs and data split ratio. The data split ratio of 7:2:1, 4:3:2:1, and 4:2:2:1:1 for 3, 4, and 5 UAVs, respectively, records the lowest loss values. The green curves, which represent the slight data-imbalance split ratios, undoubtedly have the lowest loss rates. 

This phenomenon can be explained via the problem of internal covariate shift. The solution may reach a local optimum instead of a global optimum during the training process since the batch sizes are the same. Yet, with a slight data-imbalance, there is a variation of data sizes; for example, a UAV with a large amount of data participates in training. This will shift the distribution of inputs flowing through the network and center it around the same mean and standard deviation. Hence in a three UAV setting, the data split ratio where one UAV holds 70\,\% of the data, another UAV contains 20\,\% of the data, and the last UAV possesses 10\,\% of the data, has the best performance.

\subsection{Discussion of Split Learning}
We conclude that the number of clients in split learning can be selected depending on the scale of the experiment. For example, if the surveillance UAV is to be deployed to cover a relatively large area, then at least five surveillance UAVs should be selected. There is no discrepancy in the classification performance as the number of split learning participants vary. On the other hand, if a UAV is to watch over a small area of land, then three surveillance UAVs can be selected without the comprising performance degradation. 

\begin{figure}[t]
\center
    \subfloat[\centering Raw UAV CCTV data]{
    \includegraphics[width=0.4\linewidth]{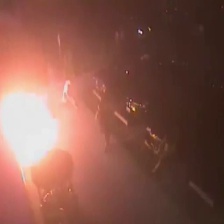}
    }
    \qquad
    \subfloat[\centering Distorted UAV CCTV data]{
    \includegraphics[width=0.4\linewidth]{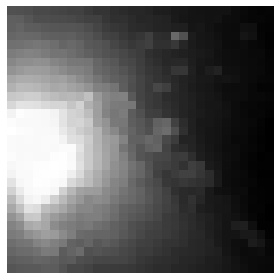}
    }
    \caption{Image (a) is the raw data of an ignition captured by the surveillance UAV. Image (b) is the output from the privacy-preserving layer. }
    \label{img:feature}
\end{figure}

Furthermore, the extent to which the privacy-preserving layer protects the original image is shown in Figure~\ref{img:feature}. Image (a) is an example of the original image captured by a surveillance UAV. After this data runs through its local first hidden layer, the output is a highly distorted image depicted in image (b). This image (b) is the only data that leaves the original host, which is the UAV and is transmitted to a server through an external network. The resulting image is hardly retraceable once the raw image is processed through its first and only hidden layer. 

As consistently repeated in this paper, applying split learning in this surveillance UAV environment creates the benefit of decreased data transmission power. Compared to fully local computation, offloading the feature map to the server will reduce the computation time and the data transmission time.

\section{Conclusions}\label{sec:sec5}

Urban air mobility (UAM) is leading the world's future mobility solution, and CCTV cameras attached to aerial vehicles are next in line. Surveillance UAVs can act as an intelligent and efficient method to fight off larceny, criminal activity and track extreme weather conditions. This paper created a scenario where multiple surveillance UAVs monitor the neighborhood for any signs of fire. 
Nevertheless, these captured images may contain private personal information such as people's faces or car plate numbers. To detect whether the fire has emerged in the streets or not, we introduce split learning into this scenario. Experiments are conducted to test the effect of increasing the number of UAVs participating in the training of CNN in a split learning environment. To examine the impact data-imbalance has on split learning, we conducted studies while varying the data split ratio between the UAVs. Through these precise experimental procedures, we came to the conclusion that the number of participants in split learning does not affect the classification rate. Yet, the slight data-imbalance should be met to reach optimal performance. Mobile devices such as UAVs have limited resources. Thereby, offloading its privacy protected data to a nearby server reduces the computational load the UAV has to endure.

\section*{Acknowledgment}

This research was supported by the National Research Foundation of Korea (NRF) (2021R1A4A1030775 and 2019M3E3A1084054). Joongheon Kim and Jae-Hyun Kim are the corresponding authors of this paper.

\bibliographystyle{IEEEtran}
\bibliography{ref_arvr,ref_aimlab,ref_uav}
\end{document}